\pdfoutput=1

\documentclass[11pt]{article}

\usepackage[final]{acl}

\usepackage{times}
\usepackage{latexsym}
\usepackage{enumitem}
\DeclareMathAlphabet\mathbfcal{OMS}{cmsy}{b}{n}
\def\@fnsymbol#1{\ensuremath{\ifcase#1\or \dagger\or *\or \ddagger\or
   \mathsection\or \mathparagraph\or \|\or **\or \dagger\dagger
   \or \ddagger\ddagger \else\@ctrerr\fi}}

\usepackage[T1]{fontenc}

\usepackage[utf8]{inputenc}

\usepackage{microtype}

\usepackage{inconsolata}

\usepackage{caption}
\usepackage{stackengine}
\usepackage{graphicx}
\usepackage{multirow}
\usepackage{booktabs}
\usepackage{enumitem}
\usepackage{amssymb}
\usepackage{pifont}
\usepackage{tabularx}

\newcommand{\cmark}{\ding{51}}%
\newcommand{\xmark}{\ding{55}}%

\usepackage{csquotes}

\title{\textit{SHARE}: Shared Memory-Aware Open-Domain Long-Term \\ Dialogue Dataset Constructed from Movie Script}

\author{%
  Eunwon Kim$^{1,}$\thanks{\; Equal contribution}\;\; 
  Chanho Park$^{1,}$\footnotemark[1]\;\; 
  Buru Chang$^{2,}$\thanks{\; Corresponding author}\\
  \textsuperscript{1}Sogang University\;\; 
  \textsuperscript{2}Korea University\\
  \texttt{\{eun1k, cksgh0984\}@sogang.ac.kr} \;\;  
  \texttt{buru\_chang@korea.ac.kr}
}

\begin{document}
\maketitle

\begin{abstract}\label{sec:0_abstract}
\textit{Shared memories} between two individuals strengthen their bond and are crucial for facilitating their ongoing conversations. 
This study aims to make long-term dialogue more engaging by leveraging these shared memories. 
To this end, we introduce a new long-term dialogue dataset named \textit{SHARE}, constructed from movie scripts, which are a rich source of shared memories among various relationships. 
Our dialogue dataset contains the summaries of persona information and events of two individuals, as explicitly revealed in their conversation, along with implicitly extractable shared memories.
We also introduce \textit{EPISODE}, a long-term dialogue framework based on \textit{SHARE} that utilizes shared experiences between individuals. 
Through experiments using \textit{SHARE}, we demonstrate that shared memories between two individuals make long-term dialogues more engaging and sustainable, and that \textit{EPISODE} effectively manages shared memories during dialogue.
Our dataset and code are available at \url{https://github.com/e1kim/SHARE}.
\end{abstract}
\section{Introduction}\label{sec:1_introduction}

\begin{displayquote}
\textit{There is nothing that can equal the treasure of so many shared memories. }
\vspace{-1em}
\begin{flushright}
--- Antoine de Saint-Exupéry
\end{flushright}
\end{displayquote}

\textit{Memory} in dialogue plays a crucial role in building relationships and rapport between individuals, and facilitating the ongoing conversation~\cite{alea2003you,nelson2003self}.
Recent studies~\cite{xu2022beyond} have proposed methods that aim for more engaging long-term dialogue by utilizing such memories. 
In particular, these methods involve summarizing and storing information about a persona~\cite{xu2022long,kwon2023and,kim2024commonsense} or personal events~\cite{bae2022keep,wang2023recursively} from the dialogue history as a memory, and incorporating this information into the response generation.

\begin{figure}[t]
  \centering
  \includegraphics[width=\columnwidth]{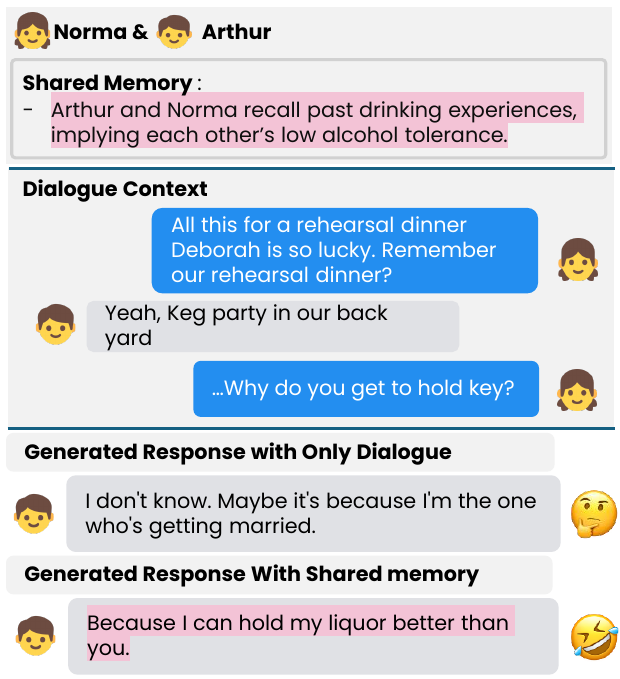}
  \caption{
Conversations between close individuals often revolve around their shared memories. Leveraging these memories to generate responses can better reflect their relationship and make the dialogue more engaging.}
  \label{fig:1_long-term_dialouge_example}
  \vspace*{-1em}
\end{figure}
However, such methods focus only on the utilization of persona information (\textit{e.g}, I'm a fan of K-Pop.) or short-term events (\textit{e.g}., doctor's appointment) for long-term dialogue. 
Although these approaches are effective in conversations between individuals getting to know each other or in everyday chit-chat, it fails to fully address conversations between individuals with a wealth of shared memories, akin to old friends.
For example, in real-world interactions between close friends who know each other well, conversations often revolve around shared memories, as shown in Figure~\ref{fig:1_long-term_dialouge_example}. 
This type of conversation is a very valuable for making long-term dialogue more engaging; however, there is a lack of research addressing it.

In this study, we introduce a new open-domain long-term dialogue dataset, \textit{SHARE}, which includes not only personas and personal event information but also information about memories shared between two speakers. 
Collecting dialogue data that contain shared memory information poses the following challenges: 
First, as in previous studies~\cite{xu2022long,kwon2023and}, crowdsourced dialogue collection requires manually creating situations for crowd-workers to role-play, leading to high costs.
Second, as in previous studies~\cite{jang2023conversation,maharana2024evaluating}, attempts to extract events using large language models (LLMs) are primarily designed to summarize explicitly stated events within dialogues. 
Consequently, they often overlook implicitly inferred events that can be captured from the conversation context.
To address these challenges, we compile long-term dialogue data from movie scripts. 
Movie scripts use dialogue to depict characters, their relationships, and explicit events, as well as to convey shared memories that are not directly revealed in scenes~\cite{kozloff2000overhearing}.
Thus, movie scripts server as a valuable source for collecting dialogue data that reflects shared memories between two individuals.
Initially, we use a movie script parser~\cite{baruah2023character} to format the movie scripts into dialogues. 
Subsequently, we utilize an LLM with a carefully designed prompt to summarize information about each speaker’s profile and short-term events, as well as to extract shared memories that are implicitly conveyed within the dialogues.
We then map this information to the corresponding utterances for further annotation.

Furthermore, based on the new dataset \textit{SHARE}, we present a long-term dialogue framework called \textit{EPISODE}. 
This framework summarizes and manages persona information, personal events, and shared memories between speakers from previous dialogue sessions, and then applies this information in long-term dialogue. 
To verify the effectiveness of using shared memories between two individuals in long-term dialogues, we conduct extensive experiments and analyses on the new dialogue dataset \textit{SHARE} with our dialogue framework \textit{EPISODE}. 
Experimental results show that the shared memories included in our dataset are effective for long-term dialogues in terms of coherence, engagingness, and reflectiveness of their relationship.

The contributions of this work are as follows:
\begin{itemize}
    \item We introduce a new long-term dialogue dataset, named \textit{SHARE}, constructed from movie scripts. 
    The new dataset includes not only persona information and personal events for each speaker but also shared memories between speakers.
    \item We present a long-term dialogue framework, \textit{EPISODE}, which summarizes persona information and personal events and extracts shared memories from dialogues. Then, the framework incorporates them into response generation for long-term dialogue.
    \item We conduct extensive experiments to verify the effectiveness of leveraging shared memories between two speakers, focusing on coherence, engagement, and reflection of their relationship in long-term dialogue. 
\end{itemize}

\section{Related Work}\label{sec:2_related_work}
\subsection{Long-term Dialogue Dataset}\label{subsec:2_1_dataset}
Some long-term dialogue datasets have recently been proposed to facilitate long-term dialogue by incorporating longer context.
\textit{MSC}~\cite{xu2022beyond} is the first persona-aware multi-session dialogue dataset, containing conversations between crowd-workers with hand-crafted persona information. 
\textit{GapChat}~\cite{zhang2023mind} includes the time intervals between sessions in the dataset, allowing the model to perceive time when conducting conversations.
\textit{CONVERSATION CHRONICLES}~\cite{jang2023conversation} is a machine-generated dataset that features relationships among speakers in the context of long-term dialogues. 
\textit{LoCoMo}~\cite{maharana2024evaluating} is a multi-modal long-term dialogue dataset created by using \texttt{GPT-3.5-turbo}, including visual images.

These datasets primarily focus on long-term dialogues based on relatively short-term events or static personas which are explicitly revealed in the dialogue.
In contrast, our dataset, \textit{SHARE}, additionally considers the longstanding and invaluable shared memories and experiences between speakers to foster their ongoing conversations.

\begin{table*}[!t]
\centering
\small
\resizebox{\textwidth}{!}{%
\begin{tabular}{l|cccccc}
\toprule 
\textbf{Dataset} & \textbf{Persona}&\textbf{Personal Event}& \textbf{Mutual Event} & \textbf{Shared Memory} & \textbf{Speakers}& \textbf{Language}  \\\midrule 
\textit{MSC}~\cite{xu2022beyond} & \cmark & \cmark & \xmark & \xmark & {P-P} & {EN} \\
\textit{LoCoMo}~\cite{maharana2024evaluating} & \cmark & \cmark & \xmark & \xmark & {B-B} & {EN}  \\     
\multicolumn{1}{l|}{\textit{DuLeMon}~\cite{xu2022long}} & \cmark & \xmark & \xmark & \xmark & {P-P} & {CH} \\     
\multicolumn{1}{l|}{\textit{PerLTQA}~\cite{du2024perltqa}} & \cmark & \cmark &  \cmark& \xmark & {B-B} & {EN,CH} \\    
\multicolumn{1}{l|}{\textit{RealPersona-Chat}~\cite{yamashita2023realpersonachat}} & \cmark & \cmark  & \xmark & \xmark & {P-P}  & {JP}\\
\multicolumn{1}{l|}{\textit{SODA}~\cite{kim2022soda}} & \cmark & \cmark & \cmark & \xmark & {B-B} & {EN}\\     
\multicolumn{1}{l|}{\textit{Dolphin}~\cite{chen2024compress}} & \xmark & \cmark & \cmark &  \xmark & {B-B} & {CH} \\ 
\multicolumn{1}{l|}{\textit{CONVERSATION CHRONICLES}~\cite{jang2023conversation}} & \xmark & \cmark & \cmark & \xmark & {B-B} & {EN} \\ 
\multicolumn{1}{l|}{\textit{MSPD}~\cite{kwon2023and}} & \cmark & \cmark & \xmark & \xmark& {P-P} & {KO} \\ 
\multicolumn{1}{l|}{\textit{GapChat}~\cite{zhang2023mind}} &\xmark & \cmark & \xmark& \xmark& {P-P} & {EN} \\ 

\midrule
\multicolumn{1}{l|}{\textbf{\textit{SHARE} (ours)}} & \cmark & \cmark & \cmark & \cmark & P-P & EN\\
\bottomrule
\end{tabular}%
}
\caption{Comparison of long-term dialogue datasets. P-P, P-B, and B-B indicate the types of interactions between person-person, person-bot, and bot-bot, respectively.}
\label{tab:1_dataset_comparison}
  \vspace*{-1em}
\end{table*}

\subsection{Long-term Dialogue Method}\label{subsec:2_2_method}
Effective management of dialogue memory is crucial for long-term conversations since the dialogue model cannot remember all the conversation history with the user.
One approach to efficient management is to develop a mechanism for updating memory~\cite{bae2022keep} and managing memory by taking structured notes during the conversation~\cite{lu2023memochat}.
\textit{COMEDY} ~\cite{chen2024compress} generates dialogue by compressing all sessions, instead of using a retriever.
\textit{MemoryBank}~\cite{zhong2024memorybank} leverages the Ebbinghaus Forgetting Curve~\cite{ebbinghaus1964memory} to develop a long-term memory mechanism, enabling LLMs to retain memories similarly to humans.

\section{Dataset Collection}\label{sec:3_dataset_collection}
In this section, we describe the new long-term dialogue dataset, \textit{SHARE}.
Table~\ref{tab:1_dataset_comparison} summarizes the comparison of existing long-term dialogue datasets and our dataset.
All details of the data collection process are in the Appendix~\ref{subsec:a_1_dataset_collection}.

\noindent
\textbf{Motivation.}
In movies, dialogue serves as the most crucial mechanism for advancing the narrative. 
Dialogues describe the characters and their relationships, as well as the unfolding events between them.
Notably, dialogue plays a pivotal role in conveying information about events between characters that are not explicitly shown in scenes~\cite{kozloff2000overhearing}.
Consequently, we regard film dialogue as an excellent source for long-term dialogue data that reflects not only the personal information of speakers and the events between them but also shared memories that are not explicitly revealed.

\noindent
\textbf{Data source.}
We have collected a total of 1201 movie scripts from IMSDB,\footnote{\url{https://imsdb.com/}} DailyScript,\footnote{\url{https://www.dailyscript.com/movie.html}}
and Simply Scripts.\footnote{\url{https://www.simplyscripts.com/}}
The collected scripts, encompassing a variety of genres including romance, comedy, and action, offer the advantage of presenting a diverse array of characters, relationships, and events.

\noindent
\textbf{Movie script preprocessing.}
Movie scripts include dialogues between characters and non-dialogue elements such as scene descriptions and action directives. 
Therefore, we utilize a movie script parser~\cite{baruah2023character} to structure the collected scripts into dialogues.

In this paper, we focus on long-term dialogue between two individuals rather than multi-party interactions. 
To achieve this, each scene featuring the two characters is treated as a single session. 
We then organize sequences of these sessions into an episode between the two characters. 
Only episodes in which the characters engage in three or more sessions are included in the dataset.

\noindent
\textbf{Information extraction.}
To construct a long-term dialogue dataset, we extract the following information from the collected dialogues using \texttt{GPT-4}\footnote{\url{https://platform.openai.com/docs/models}}:
\begin{itemize}[noitemsep,leftmargin=1em]
\item \textit{Persona} information captures essential characteristics, including personality and interests.
\item \textit{Personal event} information covers transient details like current health conditions.
\item \textit{Mutual event} represents key events between two individuals that can be explicitly inferred from the current session, unfolding between them in real-time during the conversation.
Over time, they evolve into shared memories.
\item \textit{Shared memory} refers to past events that two individuals have previously shared before the current session.
We capture these events that can be implicitly inferred from the dialogue.
\end{itemize}

\begin{table}[!t]
\centering
\small
\resizebox{\columnwidth}{!}{%
\begin{tabular}{@{}llc@{}}
\toprule
\multicolumn{3}{l}{Dataset Statistics}\\\midrule
\multicolumn{1}{l}{\multirow{5}{*}{General}}&\multicolumn{1}{l}{\# of episodes} & \multicolumn{1}{c}{3,216}\\
& \multicolumn{1}{l}{\# of sessions} & \multicolumn{1}{c}{17,679}\\
& \multicolumn{1}{l}{\# of utterances} & \multicolumn{1}{c}{119,087}\\
& \multicolumn{1}{l}{Avg. sessions per episode} & \multicolumn{1}{c}{5.50}\\
& \multicolumn{1}{l}{Avg. utterances per sessions} & \multicolumn{1}{c}{6.74}\\\midrule
\multicolumn{1}{l}{\multirow{5}{*}{Annotation}}&\# of persona info. & \multicolumn{1}{c}{80,495}\\
& \multicolumn{1}{l}{\# of personal event} & \multicolumn{1}{c}{12,436}\\
& \multicolumn{1}{l}{\# of mutual event} & \multicolumn{1}{c}{20,703}\\
& \multicolumn{1}{l}{\# of shared memory} & \multicolumn{1}{c}{4,206}\\
& \multicolumn{1}{l}{\% of episodes with shared memory} & \multicolumn{1}{c}{61.57}\\
\bottomrule
\end{tabular}%
}
\caption{The statistics of \textit{SHARE}.}
\label{tab:2_dataset_statistics}
\vspace*{-1em}
\end{table}

\begin{figure*}[t]
  \centering
  \includegraphics[width=\textwidth]{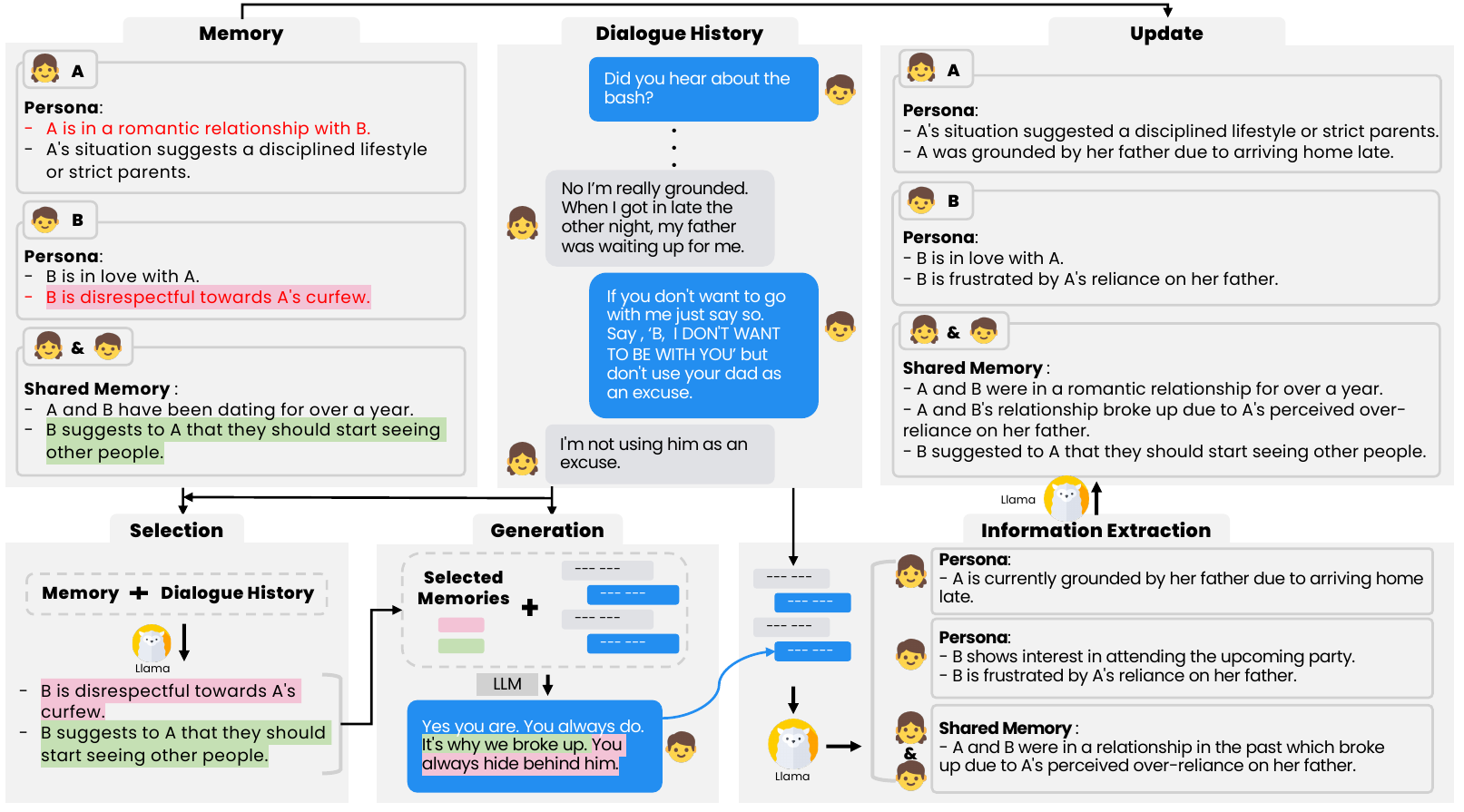}
  \caption{An architecture of \textit{EPISODE}. Our framework selects memories from the memory set based on the current dialogue history to generate responses. When the session ends, it extracts the information that needs to be stored and updates the existing memory set.}
  \label{fig:2_episode_architecture}
  \vspace*{-1em}
\end{figure*}

\noindent
\textbf{Annotation.}
To understand which information from the memory set is used to generate each utterance, we conduct annotation for each one based on the dialogue history.
This process is essential for our dialogue framework to select appropriate information when generating subsequent dialogues.
During the annotation process, we employ \texttt{GPT-3.5-turbo} to label each utterance based on the memory set information used to generate it.
For example, the utterance `\textit{We had so much fun at last year’s festival, this year will be great too!}' is annotated with `\texttt{(Shared memories)} \textit{Speaker1 and Speaker2 enjoyed last year's festival together}' from the memory set.

Finally, the statistics of the new dataset are summarized in Table~\ref{tab:2_dataset_statistics}.
We want to emphasize that 61.57\% of episodes feature at least one shared memory in their conversations.
This shows the significance of shared memory in maintaining meaningful dialogues between individuals who have known each other for a long time.
Interestingly, our dialogue dataset includes a diverse range of conversational styles, not only from everyday conversations but also from various genres, including fantasy. 
With dialogue models being increasingly applied beyond everyday conversations, such as in games~\cite{akoury2023framework} and films~\cite{chen2023large}, our dataset is a key resource for improving their adaptability across diverse dialogue styles.

\section{Dialogue Framework}\label{sec:4_dialogue_framework}
In this section, we describe the task of long-term dialogue and our dialogue framework, \textit{EPISODE}.
Figure~\ref{fig:2_episode_architecture} depicts the architecture of \textit{EPISODE}.
\textit{EPISODE} consists of a response generation procedure, including memory selection and response generation, and a memory management procedure, including information extraction and memory update, conducted after the session ends. 
Detailed information on prompts and data collection for these components is provided in Appendix~\ref{subsec:a_2_training_details}.

\subsection{Task Definition}\label{subsec:4_1_task_definition}
Let $\mathbf{e}_{(u,v)}$ = [$\mathbf{s}^{(1)}, \mathbf{s}^{(2)}, \cdots$] be an episode of two individuals $u$ and $v$ where each session $\mathbf{s}^{(i)}$ = [$\mathbf{u}_1, \mathbf{v}_1, \mathbf{u}_2, \mathbf{v}_2, \cdots, \mathbf{u}_n, \mathbf{v}_n$] is a sequence of their utterances $\mathbf{u}, \mathbf{v}$.
The dialogue model $f(\cdot)$ generates a response $\tilde{\mathbf{v}}_n$ for the given context $\mathbf{c}$ = [$\mathbf{u}_1, \mathbf{v}_1,\cdots,\mathbf{u}_n$] and memory set $\mathcal{M}_{(u,v)}$ of the individuals as follows: $\tilde{\mathbf{v}}_n = f(\mathbf{c}, \mathcal{M}_{(u,v)})$.
In this paper, the memory set consists of persona information $\mathcal{P}_u, \mathcal{P}_v$ and personal events $\mathcal{E}_u, \mathcal{E}_v$ for each individual, and shared memories $\mathcal{S}_{(u,v)}$ between them, as $\mathcal{M}_{(u,v)}=\{\mathcal{P}_u, \mathcal{P}_v, \mathcal{E}_u, \mathcal{E}_v, \mathcal{S}_{(u,v)}\}$.
Note that all types of memory used in our study are extracted from conversations between two individuals. 
Therefore, we assume that both individuals have access to and can utilize these memories.

\subsection{Response Generation}
\textbf{Memory selection.}
As long-term dialogue continues, the amount of persona information, personal events, and shared memories between the two individuals gradually increases. 
Using this expanding memory at each response generation stage is impractical. 
Therefore, we adopt a method that selects relevant memories for response generation based on the current context at each stage. 
This approach allows us to utilize memories from earlier conversations, supporting long-term dialogue. 
First, we design a prompt for memory selection and construct training data for the memory selection model.
To support dialogue situations that do not require specific memories, such as casual chitchat, we add `\texttt{Everyday Language}' to the candidate pool where memories can be selected. 
The prompt instructs the model to select `\texttt{Everyday Language}' when no suitable memory is available for the current context.
Then, using Llama-3-8B~\cite{dubey2024llama}, we train a memory selector to identify useful memories $\mathbf{m}$ from the current memory set $\mathcal{M}_{(u,v)}$ for response generation based on the given context $\mathbf{c}$.

\noindent
\textbf{Response generation.}
We use Gemma~\cite{team2024gemma} and Llama-3-8B~\cite{dubey2024llama} as the dialogue model $f(\cdot)$ to generate responses $\tilde{\mathbf{v}}_n$ appropriate to the given context $\mathbf{c}$ by utilizing the memories $\mathbf{m}$ selected in the previous step: $\tilde{\mathbf{v}}_n=f(\mathbf{c},\mathbf{m})$.
To enable the dialogue model to incorporate the selected memories into response generation, we design specific prompts and collect training data based on these prompts to train the LLM.
From this, the dialogue model generates personalized and consistent dialogues by providing responses that align with the memory while also considering the conversation history.
\subsection{Memory Management}
When the current session ends, \textit{EPISODE} manages memory asynchronously. 
This approach prevents any additional latency in response generation from the user’s perspective.

\noindent
\textbf{Information extraction.}
At the end of each dialogue session, an information extraction process is triggered to identify distinct and memorable details that may be useful for future interactions, following the approach used in the SHARE dataset. 
The extracted information is then incorporated into subsequent sessions. To minimize reliance on external APIs, we train Llama-3-8B on information extraction data collected during the construction of SHARE and utilize it as our information extractor.

\noindent
\textbf{Memory update.}
Storing all information generated in each session significantly increases the memory size over time, raising computation costs and making it difficult to select memory relevant to the current context. 
Additionally, persona information, personal events, and mutual events between the two individuals are continuously updated or newly added. 
Therefore, if the memory is not regularly updated and refined, inappropriate memories may be selected, leading to a decline in the quality of future dialogues~\cite{bae2022keep}. 
To address this, we continuously refine and manage memory using the following strategies:
\begin{itemize}[noitemsep,leftmargin=1em]
    \item \textbf{Accumulate extracted information}: When the extracted information is independent of existing memory, it is directly added to the memory. If the information involves a mutual event, it is stored as shared memory.

    \item \textbf{Update logically sequential information}: If the extracted information is sequentially or causally related to the existing memory, the memory is updated to reflect this relationship.

    \item \textbf{Update logically conflicting information}: When the extracted information conflicts with or contradicts the existing memory, the memory is updated to reflect these changes, ensuring alignment with the new information.

    \item \textbf{Deduplicate information}: If newly extracted information duplicates existing memory, the memory is not updated to maintain efficiency.
\end{itemize}

\begin{table*}[!t]
\centering
\small
\resizebox{\textwidth}{!}{
\begin{tabular}{@{}l|l|cccccccc@{}}
\toprule
Backbone& Method&BLEU-3/4$\uparrow$&ROUGE-1/2$\uparrow$&ROUGEL$\uparrow$&BertSim$\uparrow$&Dist-1/2$\uparrow$&ppl$\downarrow$\\\midrule
\multirow{5}{*}{Gemma (2B)}&\textit{Gemma (zero-shot)} &{0.0120}/\underline{0.0096} &\underline{0.0999}/0.0249&\underline{0.0957}&\underline{0.8648}&\textbf{0.6702/0.8718}&24.806\\
& \textit{Gemma}+\textit{SHARE {\footnotesize w/o memory}}  &\underline{0.0124}/{0.0089} &0.0991/\underline{0.0262}&{0.0940}&{0.8589}&{0.6462/0.8619}&6.7083 \\

& \textit{Gemma}+\textit{SHARE {\footnotesize w/ predicted individual memory}}  &0.0073/0.0059 &0.0808/0.0159&0.0759&0.8526&0.6365/0.8498&\underline{4.1836} \\

& \textit{Gemma}+\textit{SHARE {\footnotesize w/ predicted memory}}  &\textbf{0.0134/0.0110} &\textbf{0.1129/0.0306}&\textbf{0.1060}&\textbf{0.8682}&\underline{0.6605}/\underline{0.8696}&\textbf{3.7487} \\

\cmidrule{2-8}
& \textit{Gemma}+\textit{SHARE {\footnotesize w/ gold memory}}  &0.0230/0.0189 &0.1468/0.0530&0.1377&0.8674&0.6636/0.8546&3.7911 \\\midrule
\multirow{5}{*}{LLaMA-3.1-Instruct (8B)}&\textit{LLaMA (zero-shot)}&{0.0122}/0.0099 &{0.0997}/{0.0213}&{0.0923}&{0.8474}&{0.5458}/\textbf{0.8470}&23.942 \\
& \textit{LLaMA}+\textit{SHARE {\footnotesize w/o memory}}  &\underline{0.0168}/\underline{0.0135} &\underline{0.1146}/\underline{0.0329}&\underline{0.1085}&\underline{0.8592}&\textbf{0.6372}/\underline{0.8432}&5.8660 \\

&
\textit{LLaMA}+\textit{SHARE {\footnotesize w/ predicted individual memory}}  &0.0145/0.0116&0.0988/0.0272&0.0927&0.7967&0.5620/0.7670&\underline{3.3393} \\

& \textit{LLaMA}+\textit{SHARE {\footnotesize w/ predicted memory}}  &\textbf{0.0267/0.0200}&\textbf{0.1392}/\textbf{0.0508}&\textbf{0.1290}&\textbf{0.8632}&\underline{0.5676}/0.8179&\textbf{3.2409} \\
\cmidrule{2-8}
& \textit{LLaMA}+\textit{SHARE {\footnotesize w/ gold memory}}  &0.0500/0.0377&0.2205/0.1000&0.2040&0.8806&0.6171/0.8389&3.1277
\\\bottomrule
\end{tabular}%
}
\caption{Comparison of automated evaluation metric result across various systems}
\label{tab:3_experimental_results}
\end{table*}

\begin{table*}[!t]
\centering
\small
\resizebox{\textwidth}{!}{
\begin{tabular}{@{}l|l|cccc|cccc|cccc@{}}
\toprule
\multirow{2}{*}{Backbone}& \multirow{2}{*}{Method}&\multicolumn{4}{c|}{Session 4}&\multicolumn{4}{c|}{Session 5}&\multicolumn{4}{c}{Session 6}\\
&&Coh.&Eng.&Clo.&Ref.&Coh.&Eng.&Clo.&Ref.&Coh.&Eng.&Clo.&Ref.
\\\midrule
\multirow{6}{*}{Gemma (2B)}
& \textit{SHARE} (w/o shared memory) &\underline{1.8104}&1.5812&\underline{1.2604}&&1.9188&1.5854&\underline{1.3167}&&1.9542&1.6958&1.2479\\\cmidrule{2-14}

& \textit{SHARE} + \textit{ACCUMULATION} &1.7292&\textbf{1.7458}&1.2083&0.8125&\underline{1.9854}&\textbf{1.7711}&1.3104&0.9625&\underline{2.0312}&\textbf{1.8250}&\textbf{1.3354}&1.0729 \\ \

& \textit{SHARE} + \textit{COMEDY} &1.0246&1.1896&1.1667&&0.9979&1.1521&1.0604&&1.0854&1.1500&1.0479\\

& \textit{SHARE} + \textit{LLM-Rsum}&1.5021&1.0813&0.9042&&1.3979&1.1042&0.7521&&1.6146&1.1792&0.9542\\

\cmidrule{2-14}
& \textit{SHARE} + \textit{EPISODE} &\textbf{2.0187}&\underline{1.6646}&\textbf{1.2854}&\textbf{1.0542}&\textbf{2.1729}&\underline{1.7708}&\textbf{1.3271}&\textbf{1.1812}&\textbf{2.0500}&\underline{1.7042}&\underline{1.3057}&\textbf{1.1375} \\
\cmidrule{1-14}

\multirow{6}{*}{LLaMA-3.1-Instruct (8B)}
& \textit{SHARE} (w/o shared memory) &2.5000&2.2438&1.5979&&2.5375&\underline{2.2875}&1.7021&&2.5979&\underline{2.3538}&1.6771
\\\cmidrule{2-14}
& \textit{SHARE} + \textit{ACCUMULATION} &\underline{2.6125} &\underline{2.2875}&\textbf{1.7354}&1.0292&\textbf{2.5833}&2.2687&\underline{1.7167}&1.2583&2.5958&2.3125&\underline{1.7271}&1.3937 \\ 

& \textit{SHARE} + \textit{COMEDY} &2.0104&1.2771&0.9854&&2.0041&1.3000&0.9833&&2.0625&1.2521&1.0042\\

& \textit{SHARE} + \textit{LLM-Rsum} &2.5166&1.7646&1.2792&&\underline{2.5709}&1.7146&1.1583&&\textbf{2.6208}&1.7063&1.2688\\

\cmidrule{2-14}

& \textit{SHARE} + \textit{EPISODE} &\textbf{2.6708}&\textbf{2.3313}&\underline{1.6896}&\textbf{1.4458}&2.5542&\textbf{2.3750}&\textbf{1.7313}&\textbf{1.5833}&\underline{2.6042}&\textbf{2.3625}&\textbf{1.7583}&\textbf{1.7604} \\\bottomrule
\end{tabular}%
}
\caption{Experimental results of multi-session \texttt{GPT-4o} evaluation.}
\label{tab:4_experimental_results_multisession}
\vspace*{-1em}
\end{table*}

\section{Experiments}\label{sec:5_experiments}
\subsection{Experimental Setup}\label{subsec:5_1_experimental_setup}


\noindent
\textbf{Backbone models.}
We employ state-of-the-art generative models, Gemma (2B)~\cite{team2024gemma} and Llama-3.1-Instruct (8B)~\cite{dubey2024llama}, as the backbone model of response generation.

\noindent
\textbf{Evaluation metrics.}
To assess the quality of the generated responses for each session, we conduct an automatic evaluation using a comprehensive set of metrics, including BLEU-3/4~\cite{papineni2002bleu}, ROUGE (ROUGE-1/2/L)~\cite{lin2004rouge}, BERTscore~\cite{zhang2019bertscore}, Distinct-1/2 (Dist)~\cite{li2015diversity} and PPL. 
These metrics offer a thorough analysis of the model's performance and the diversity of the generated responses.

To verify that using shared memory makes long-term dialogues more engaging, we further conduct a \texttt{GPT-4o} assisted evaluation based on the following criteria: \textbf{Coherence (Coh.)}: measures the consistency and logical connection of responses within a session. \textbf{Engagingness (Eng.)}: assesses the speaker's ability to maintain the annotator's interest for a long-term dialogue. \textbf{Closeness (Clo.)}: determines how well the two participants know each other, considering the depth of their familiarity and understanding. \textbf{Reflectiveness (Ref.)}: evaluates how well the dialogue reflects the relationship indicated in the memory set.

\noindent
\textbf{Implementation details.}
We partition the SHARE dataset into an 8:1:1 ratio based on episodes, allocating the splits for training, validation, and testing in our long-term dialogue framework, EPISODE.
Further details on these experiments are provided in Appendix~\ref{subsec:a_6_experiments}.

\subsection{Experimental Results}\label{subsec:5_2_experimental_results}
\textbf{Automatic evaluation.}
We evaluate the generated responses for the final utterance of all sessions in the test set.
We compare the following three versions of dialogue generation trained on \textit{SHARE} for each backbone model.

\begin{itemize}[noitemsep]
    \item 
    \textbf{\textit{SHARE {\footnotesize w/o memory}}} is trained only with the dialogue from \textit{SHARE} without using memories.
    
    \item
    \textbf{\textit{SHARE {\footnotesize w/ predicted memory}}} is trained with both the dialogue and annotated memories from \textit{SHARE}.
    During inference, it generates the responses based on the memory selection for the given dialogue context. 

    \item
    \textbf{\textit{SHARE {\footnotesize w/ predicted individual memory}}} is a baseline where each speacker independently manages their memory for response generation.
    \item 
    \textbf{\textit{SHARE {\footnotesize w/ gold memory}}} uses the gold memory to generate responses during inference.

\end{itemize}

Table~\ref{tab:3_experimental_results} shows the comparison of automatic evaluation metric results.
Training with the \textbf{\textit{SHARE}} dataset improves most performance metrics for generating the next response.
In Llama-3, the BLEU-3/4 and ROUGE scores for \textbf{\textit{SHARE {\footnotesize w/ predicted memory}}} responses are higher than those for \textbf{\textit{SHARE {\footnotesize w/ zero-shot}}} and \textbf{\textit{SHARE {\footnotesize w/o memory}}}
This reflects better performance in dialogue generation for using the \textit{SHARE} dataset, especially shared memory.
Also, \textbf{\textit{SHARE {\footnotesize w/ predicted memory}}} outperforms \textbf{\textit{SHARE {\footnotesize w/ predicted individual memory}}}, demonstrating that incorporating the interlocutor’s persona and personal event information leads to more diverse and enriched conversations compared to their removal.
In summary, these results show that \textbf{\textit{SHARE {\footnotesize w/ predicted memory}}} is a response generation model that effectively incorporates shared memory.

\noindent
\textbf{Multi-Session Evaluation.}
As the number of sessions increases, the update model efficiently organizes the memory $\mathcal{M}_{(u,v)}$.
To assess the performance of the updated model, we use a GPT-based evaluation approach.
The test set represents as $\mathbf{e}_{(u,v)} = [\mathbf{s}^{(1)}, \mathbf{s}^{(2)}, \dots,\mathbf{s}^{(6)}]$, including more than six sessions, and a total of 96 episodes are selected. 
Each session $\mathbf{s}^{(i)}$ contains the utterances of two individuals, $u$ and $v$, structured as $[\mathbf{u}_1, \mathbf{v}_1, \mathbf{u}_2, \mathbf{v}_2, \dots, \mathbf{u}_{10}, \mathbf{v}_{10}]$. 
The first two utterances are provided, and the model generates the following eight utterances, completing the session with a total of 10 sentences.
Before generating the third session, the updated model integrates the information from the first and second sessions into $\mathcal{M}_{(u,v)}$, which is then used to generate the third session.
After the session concludes, the relevant information is extracted. 
From the fourth session, the model updates its memory with data from the past two sessions to generate each new session, continuing until the sixth.

\textit{SHARE}+\textit{EPISODE} refers to our update model, which updates the memory throughout the sessions, while \textit{SHARE}+\textit{ACCUMULATE} accumulates all information as sessions progress without any updates.
In the \textit{SHARE} (w/o shared memory), $\mathcal{S}_{(u,v)}$ is excluded during memory selection.
As a baseline, we use the \textit{COMEDY} \cite{chen2024compress} model and \textit{LLM-Rsum}\cite{wang2023recursively}. 
\textit{COMEDY} analyzes previous session data to extract key details, such as significant events, user tendencies, and the nature of user-bot interactions, and then compresses them into a single memory. 
\textit{LLM-Rsum} recursively summarizes and updates stored information at the end of each session.
We evaluate the four metrics described in Section \ref{subsec:5_1_experimental_setup} on a scale from 0 to 3 points. 
More detailed information are in Appendix \ref{subsec:a_6_experiments}.

\begin{figure*}[t]
  \centering
  \includegraphics[width=\textwidth]{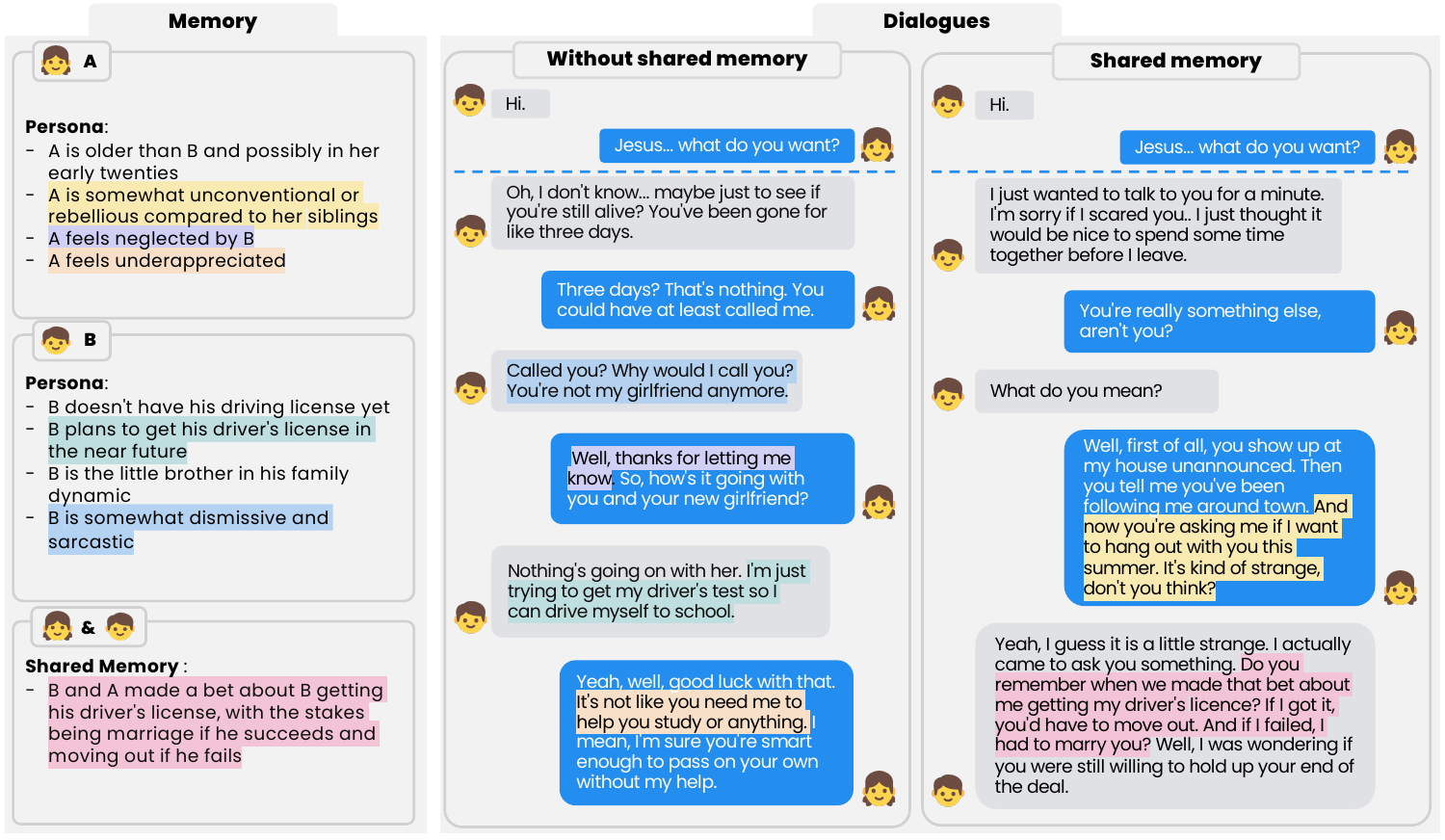}
  \caption{Examples of dialogues with and without shared memory. The first and second utterances are the same in both dialogues, with the models generating responses from the third utterance onward. The conversation on the right builds upon shared memory from previous sessions.}
  \label{fig:case_study}
  \vspace*{-1em}
\end{figure*}

Table \ref{tab:4_experimental_results_multisession} shows the evaluation of multi-session dialogues generated by the fine-tuned model. 
\textit{SHARE}+\textit{EPISODE}  model consistently outperforms others in reflectiveness.
Although the Llama-3 shows lower scores in closeness in session 4, its performance improves in subsequent sessions, eventually surpassing the \textit{Accumulate} model.
This highlights the model’s effective use of shared experiences and clear depiction of the participants’ relationship across multiple sessions.

Compared to the baselines, Gemma achieves higher metrics per session.
Since LLM-Rsum recursively summarizes and updates stored information, evaluating the significance of individual memory units becomes challenging, which in turn leads to reduced Eng scores and lower Clo performance in conversations.
The above results highlight the success of the EPISODE framework, which enhances conversational capabilities and effectively extracts both explicit and implicit information.
Also, this approach proves to be more effective than previous approaches, which are primarily designed to summarize mainly explicitly stated events in dialogues, overlooking implicit experiences.

We also perform human evaluations under the same conditions, assessing key aspects such as coherence, reflectiveness, and engagingness of the generated dialogues.
Detailed results of these evaluations are provided in Appendix \ref{subsec:a_7_human_evaluation}.

\begin{table}[!t]
\centering
\resizebox{\columnwidth}{!}{
\begin{tabular}{l|ccc}
\toprule
\textbf{Method} & \textbf{Con.} & \textbf{Ref.} & \textbf{Eng.} \\ 
\midrule
\multicolumn{1}{l|}{\textbf{Gemma (2B)}} & & &  \\
\textit{SHARE} (w/o shared memory) & 1.0708 & 2.3396 & \textbf{1.3000} \\
\textit{SHARE} + \textit{EPISODE}  & \textbf{1.1021} & \textbf{2.3438} & 1.2896 \\ 
\cmidrule{1-4}
\multicolumn{1}{l|}{\textbf{Llama-3-Instruct (8B)}} & & &  \\
\textit{SHARE} (w/o shared memory) & 2.0250 & 2.3129 & 1.6563 \\
\textit{SHARE} + \textit{EPISODE}  & \textbf{2.0729} & \textbf{2.3604} & \textbf{1.7021} \\
\bottomrule
\end{tabular}%
}
\caption{Comparison of different methods with and without shared memory based on \texttt{GPT-4o} evaluation.}
\label{tab:5_episode_evaluation}
\vspace*{-1em}
\end{table}

\noindent
\textbf{EPISODE Evaluation.}
We evaluate the entire set of sessions together, rather than a single session.
This is because, first, a good conversation should maintain consistency in the relationship between the two participants across all sessions.
For example, if two characters are close in one session but suddenly become awkward in the next, it breaks the coherence of the dialogue. 
Second, a true long-term evaluation requires analyzing the dialogue from a broader perspective.
If only one session is evaluated, important elements like references to past events that add interest could be missed.
Thus, analyzing the entire conversation is necessary to accurately assess its Consistency (Con.), Reflectiveness (Ref.), and Engagingness (Eng.).
Thus, we provide the full dialogue  \(\mathbf{e}_{(u,v)} = [\mathbf{s}^{(1)}, \mathbf{s}^{(2)}, \dots,\mathbf{s}^{(5)}]\), and evaluate the overall conversation.
In the test set, 131 episodes with more than 5 sessions are used for evaluation.
Utterances are generated using the same method as in the multi-session evaluation.
Notably, Consistency differs from coherence in that it assesses whether the relationship between the two speakers remains stable and unchanged throughout the entire episode
Detailed evaluation prompts are provided in the Appendix \ref{subsec:a_6_experiments}.

Table \ref{tab:5_episode_evaluation} presents the evaluation of the full conversation sessions (1-5 sessions) generated by our \textit{EPISODE} model.
The results indicate that applying shared memory in each model leads to better performance compared to when it is not used.
In Llama-3 and Gemma, \textit{SHARE}+\textit{EPISODE} achieves higher scores in Reflectiveness and Consistency compared to \textit{SHARE} (w/o shared memory).
This result indicates that shared memory enhances the model’s understanding of the relationship between the two participants, leading to more consistent dialogues.
While the engagingness score for the Gemma slightly decreases with shared memory, this is likely due to the inherent performance differences between the models.
The Llama-3 initially performs better, and this advantage becomes more pronounced with shared memory, further enhancing the overall quality of the conversation.
\subsection{Case Study}\label{subsec:5_3_case_study}
We examine how shared memory influences interactions between dialogue models. 
Figure \ref{fig:case_study} presents example dialogues generated by the proposed \textit{EPISODE} framework, comparing cases with and without shared memory usage.

\noindent
\textbf{Without shared memory}.
In this dialogue, the model generates responses that align with each speaker’s persona. 
However, it relies solely on persona information and does not incorporate any specific shared events or experiences between the speakers. 
Consequently, while the conversation maintains coherence, it lacks depth and a strong sense of connection, as it fails to reference unique moments that define their relationship.

\noindent
\textbf{With shared memory}.
In this dialogue, the model integrates shared experiences, exemplified by references such as, "Do you remember when we made that bet about me getting my driver’s license?" 
This demonstrates the model effectively utilizes shared memory, reinforcing past interactions between the speakers. 
By incorporating shared experiences, the conversation becomes more engaging and dynamic, naturally revealing the relationship and emotions between the participants. 
Moreover, the dialogue conveys a sense of familiarity and warmth, enhancing the authenticity of their bond and providing a more immersive user experience.
\section{Conclusion}\label{sec:6_conclusion}
In this study, we introduce \textit{SHARE}, a new long-term dialogue dataset that includes shared experiences between two participants.
This dataset, extracted from movie scripts, provides a critical foundation for the development of long-term dialogue systems.
We propose \textit{EPISODE} framework, a novel approach for long-term dialogue systems, which promotes more natural conversations.
We also utilize a novel evaluation method, GPT-Eval, to extend beyond traditional metrics and incorporate diverse evaluation criteria.
By integrating shared memory into the dialogue system, we demonstrate improvements in both reflectiveness and consistency, which lead to richer and more coherent interactions.
By making the newly collected \textit{SHARE} dataset publicly available, we aim to contribute to future research.

\section*{Limitation}\label{sec:7_limitation}
We create a dialogue dataset of two individuals with shared memories based on movie scripts.
Despite the strong performance demonstrated by the models trained on \textit{SHARE}, there are several limitations to consider.
First, since the dataset is constructed from movie scripts, the model may generate dramatic dialogues that differ from typical chit-chat.
Second, the use of imperfect machine learning models (\textit{e.g.} \texttt{GPT-4} and \texttt{GPT-3.5-turbo}) for information extraction and annotation may result in inaccuracies in this information and annotations.
Third, using multiple GPT models during dataset construction may introduce inconsistencies in the extracted information. Initially, \texttt{GPT-4} is used for information extraction, but later processes, such as memory updates and utterance mapping, employ \texttt{GPT-4o} and \texttt{GPT-3.5-turbo} to reduce costs.

\section*{Ethical Consideration}\label{sec:8_ethical_consideration}

This study construct a dialogue dataset from movie scripts that cover various topics and character relationships. 
As a result, the dataset may contain offensive language or insults targeting specific genders, races, or social groups. 
Additionally, since the dataset includes movies released over several decades, it may reflect various discriminatory content from different eras. 
Therefore, when conducting research using this dataset, it is crucial to implement safeguards to prevent inappropriate responses due to such content.

\section*{Acknowledgment}
This work was partly supported by the Institute of Information \& Communications Technology Planning \& Evaluation(IITP)-ICT Creative Consilience Program grant funded by the Korea government(MSIT)(IITP-2025-RS-2020-II201819) and the National Research Foundation of Korea (NRF) grant funded by the Korea government (MSIT) (RS-2024-00350430). Finally, this research was also supported by the Smilegate AI Center.

\bibliography{custom}

\clearpage
\appendix

\section{Appendix}\label{sec:a_appendix}
\subsection{Dataset Collection}\label{subsec:a_1_dataset_collection}
\begin{figure*}[t]
  \centering
  \includegraphics[width=\textwidth]{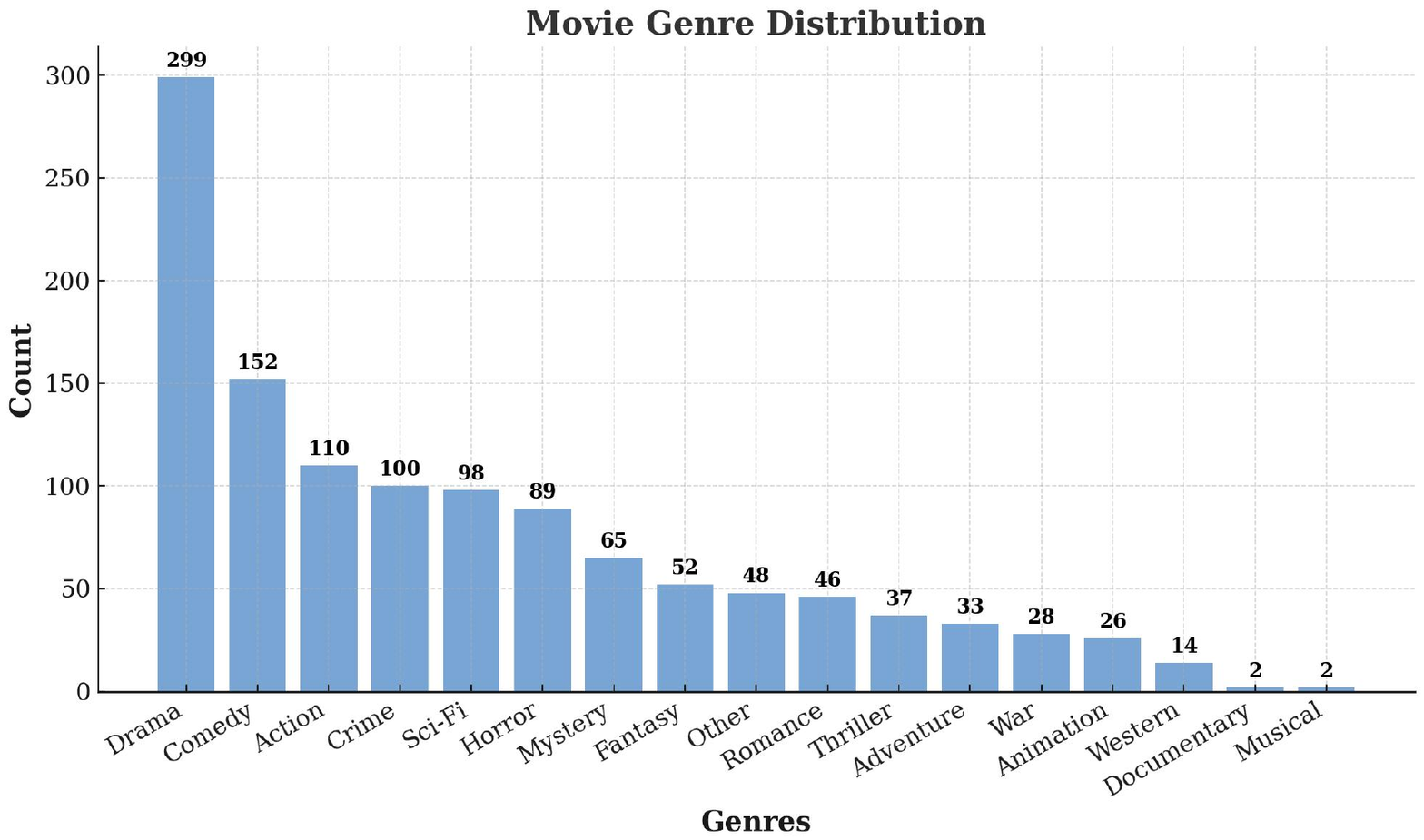}
  \caption{The number of movies in each genre category in \textit{SHARE}}
  \label{fig:movie_genre_distribution}
\end{figure*}

We use \texttt{GPT-4} to extract information from movie scripts.
The detailed prompt is provided in the Table \ref{table:prompt_information_extraction}.
Before annotation, we parse the Persona sentences because they contained too much information in a single sentence. 
The prompts for parsing and annotation are provided in Table \ref{table:prompt_parsing_information} and Table \ref{table:prompt_labeling_task}.
We use \texttt{GPT-3.5-turbo} for both parsing and annotation. In this study, we divide the dataset sessions into training(14,255), testing(1,685), and validation(1,739) sets. The distribution of dataset genres is shown in the Figure~\ref{fig:movie_genre_distribution}

\subsection{Implementation and Training Details}\label{subsec:a_2_training_details}
\noindent
\textbf{Response Generator}
We train the Response Generator model using both the Llama and Gemma models.
We train the Gemma on 2 NVIDIA RTX A5000 GPU devices, with a cosine scheduler, starting with a learning rate of $1 \times 10^{-5}$ and using 1000 warmup steps. The per-device batch size is set to 2, and gradients are accumulated over 1 step. The maximum output sequence length is 1024 tokens. The model is trained for 4 epochs. 
It takes 2 hours to train the model.
We train the Llama on 2 NVIDIA RTX A6000 GPU devices, with a cosine scheduler, starting with a learning rate of $1 \times 10^{-5}$ and using 1000 warmup steps. The per-device batch size is set to 8, and gradients are accumulated over 1 step. The maximum output sequence length is 1024 tokens. The model is trained for 4 epochs.
It takes 8 hours to train the model.

We employ Low-Rank Adaptation (LoRA)\cite{hu2021lora}, a computationally efficient tuning method, for both the LLAMA and Gemma models. LoRA reduces the trainable parameters by utilizing pairs of rank-decomposition matrices. In our experiments, we set the rank to 16

\noindent
\textbf{Information extraction module}
We use the Llama with 2 NVIDIA RTX A6000 GPUs and a batch size of 4, with a training time of 8 hours.
The maximum output sequence length is 2048 tokens.

\noindent
\textbf{Memory selection module}
We use the Llama with 2 NVIDIA RTX A6000 GPUs and a batch size of 4, with a training time of approximately 2 days.
The maximum output sequence length is 2048 tokens.

\noindent
\textbf{Memory update module}
We train the Llama on 2 NVIDIA RTX A6000 GPU devices, with a cosine scheduler, starting with a learning rate of $1 \times 10^{-5}$ and using 1000 warmup steps. 
The per-device batch size is set to 4 for both training and evaluation, and gradients are accumulated over 1 step. 
The maximum length of the output sequence is 2048 tokens.
It takes 5 hours to train the model.

\subsection{Memory selection}\label{subsec:a_3_memory_selection}
In memory selection, the most appropriate memory must be chosen to ensure contextually relevant dialogue.
To select the appropriate memory, it is necessary to consider all memories from previous sessions and all utterances from the current conversation.
To prevent the model from being influenced by the order of the memories, and to mitigate any potential bias caused by sequence ordering, the memories are shuffled before being processed.
The prompt used for memory selection is presented in Table \ref{table:prompt_memory_selection}.
We use GPT-3.5-turbo for memory annotation and compare the performance of our trained memory selection model with GPT-3.5-turbo. As shown in Table~\ref{tab:memory_selection_compare_with_gpt-3.5-turbo}, our model achieves a significantly higher F1 score, demonstrating the effectiveness of the learned memory selection model.

\begin{table}[t]
\centering
\small
\begin{tabular}{l|ccc}
\toprule
\textbf{Model} & \textbf{Recall} & \textbf{Precision} & \textbf{F1} \\
\midrule
GPT-3.5-Turbo & 0.056 & 0.033 & 0.042 \\
Ours          &\textbf{ 0.186} & \textbf{0.208} & \textbf{0.196} \\
\bottomrule
\end{tabular}
\caption{Performance comparison between our trained memory selection model and GPT-3.5-turbo}
\label{tab:memory_selection_compare_with_gpt-3.5-turbo}
\end{table}

\subsection{Information extraction}\label{subsec:a_4_information_extraction}
We conduct experiments to evaluate how similar and consistent the model’s output is compared to the actual ground truth.
The experiments are performed using the BLEU-3/4~\cite{papineni2002bleu}, ROUGE (ROUGE-1/2/L)~\cite{lin2004rouge}, and BERTscore~\cite{zhang2019bertscore} metrics, and the model was compared against \texttt{GPT-3.5-turbo}.
The results are presented in Table~\ref{tab:info_extraction}.
The prompt used during training is very similar to the one used in the information extraction part of Session \ref{sec:3_dataset_collection}.
However, in the training prompt, the phrase \texttt{“from the movie \{movie\_name\}. Use your knowledge of the characters to enhance your analysis where applicable.”} is removed. 
The details of the modified prompt can be found in Table \ref{table:prompt_information_extraction}.

\begin{table}[ht]
\centering
\begin{tabular}{l|c|c}
\toprule
\textbf{Score} & \textbf{GPT-3.5-turbo} & \textbf{Ours} \\ 
\midrule
\textbf{BertSim}   & 0.9059 & \textbf{0.9262} \\ 
\textbf{ROUGE-1} & 0.4681 & \textbf{0.5265} \\ 
\textbf{ROUGE-2} & 0.2332 & \textbf{0.3294} \\ 
\textbf{ROUGE-L} & 0.3428 & \textbf{0.4326} \\ 
\textbf{BLEU-1}   & 0.2660  & \textbf{0.3053} \\ 
\textbf{BLEU-2}   & 0.1643  & \textbf{0.2261} \\ 
\textbf{BLEU-3}   & 0.1140  & \textbf{0.1811} \\ 
\textbf{BLEU-4}   & 0.0788  & \textbf{0.1467} \\ 
\bottomrule
\end{tabular}
\caption{Comparison of Information Extraction Scores between GPT-3.5-turbo and Our Model}
\label{tab:info_extraction}
\end{table}

\subsection{Memory Update}\label{subsec:a_5_memory_update}
Storing all dialogue content over time can lead to significant overhead. 
Retaining all memories also complicates the process of selecting the most relevant memory for the conversation. 
Since individuals experience changes in their persona and events throughout life, it is natural for certain memories to evolve. 
Consequently, dynamic memories exist. 
To address this, we have developed a memory update strategy that ensures the selection of appropriate memories during conversations, even as those memories change over time.
Detailed examples of the update process and corresponding prompts are provided in Table \ref{table:memory_update_example}, Table \ref{table:prompt_memory_update}

\subsection{Experiments}\label{subsec:a_6_experiments}
Evaluating the performance of open-domain conversations is critical for advancing natural language understanding. While traditional metrics remain important, recent methods have incorporated assessments utilizing models like GPT-4 to enhance evaluation accuracy.
One such method, GPT Eval \cite{liu2023g}, leverages Chain of Thought (CoT) reasoning to evaluate a model's text generation. This technique offers greater context and structured guidance for large language models (LLMs) when assessing generated text.
For this evaluation, we select multi-session dialogues from the \textit{SHARE} dataset, each consisting of more than five sessions. From each session, two utterances are extracted to create dialogues within an \textit{EPISODE}. The experiment is conducted five times for each criterion, and the results are averaged to determine the final outcome. The GPT Eval prompt used in this process is provide in Table \ref{table:prompt_multisession_coherence}, Table \ref{table:prompt_multisession_engagingness}, Table \ref{table:prompt_multisession_closeness}, Table \ref{table:prompt_multisession_reflectiveness}, Table
\ref{table:prompt_episode_consistency}, Table 
\ref{table:prompt_episode_reflectiveness}, Table
\ref{table:prompt_episode_engagingness}.




\begin{table}[!t]
\centering
\resizebox{\columnwidth}{!}{%
\begin{tabular}{l|ccc}
\toprule
Method&Coh&Eng&Clo\\\midrule
\textit{EPISODE} {\scriptsize w/o \textit{shared memory}} &2.276&2.283&2.325\\
\textit{EPISODE} &\textbf{2.282}&\textbf{2.304}&\textbf{2.365}\\\bottomrule
\end{tabular}%
}
\caption{Experimental results of human evaluation.}
\label{tab:7_human_evaluation}

\end{table}

\subsection{Human Evaluation}\label{subsec:a_7_human_evaluation}
\begin{figure*}[t]
  \centering
  \includegraphics[width=\textwidth]{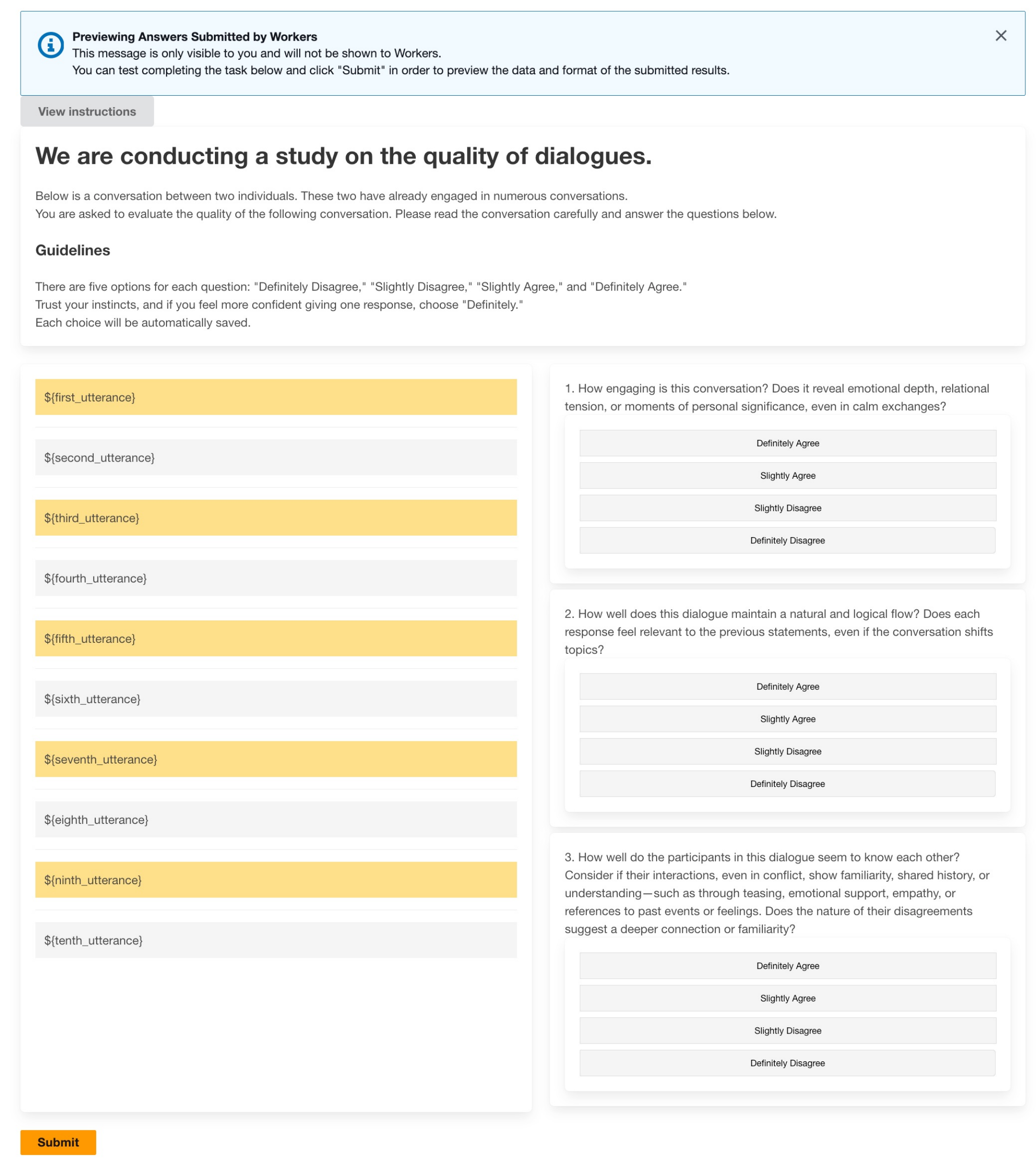}
  \caption{User interface (UI) used for human evaluation on Amazon Mechanical Turk, displaying the screen as viewed by the annotators.}
  \label{fig:4_human_evaluation_ui}
\end{figure*}

We conduct human evaluations through Amazon Mechanical Turk(AMT) to highlight the differences between \textit{SHARE} and \textit{SHARE} (w/o shared memory).
We show the interface for the evaluation in Figure \ref{fig:4_human_evaluation_ui}. 
Using a random seed of 2025, we randomly select 25 conversations from 96 test sets, which are constructed using the EPISODE Framework. 
We select annotators who are native English speakers and have a HIT Approval Rate of over 80\% for all Requesters' HITs. 
Then, we ask 20 annotators to evaluate each conversation based on the following criteria:
\begin{itemize}[noitemsep]
    \item 
    \textbf{Closeness} : In this criterion, we ask annotators to assess whether a response reflects the speakers' familiarity, shared history, or understanding, expressed through teasing, emotional support, empathy, or references to past events, even during conflicts.
    
    \item
    \textbf{Engagingness} : we ask the annotators to judge whether a response reveals emotional depth, relational tension, or moments of personal significance, even in calm exchanges.
    
    \item 
    \textbf{Coherence} : This criterion examines whether each response naturally connects to the preceding statements, even as the conversation evolves in different directions.

\end{itemize}

The evaluation takes place in the third session to maintain consistency in the testing environment.
Table \ref{tab:7_human_evaluation} provides a comparison between \textit{SHARE} and \textit{SHARE} (w/o shared memory) conversations, showing that \textit{SHARE} achieves superior results over \textit{SHARE} (w/o shared memory).
This demonstrates that conversations with shared memory are more engaging and result in more natural dialogue.



\captionsetup{justification=centering} 

\begin{table*}
    \centering
    \footnotesize
    \begin{minipage}{0.8\textwidth}
    \centering
    \resizebox{0.6\textwidth}{!}{%
    \begin{tabular}{l|l}
        \toprule
       Number  & Label\\
       \midrule
        1 & Accumulate\\ \midrule
        2 & Connect sequential/causal events\\ \midrule
        3 & Update conflicting events\\ \midrule
        4 & Deduplicate information\\ 
        \bottomrule
    \end{tabular}
    }
    \caption{Summary of memory update actions and corresponding labels}
    \label{tab:memory_update_actions}
    \end{minipage}

\vspace{0.5cm}

    \begin{minipage}{\textwidth}
    \centering 
    \resizebox{\textwidth}{!}{%
    \begin{tabular}{l|l|l}
        \toprule
        Previous memory & Current memory  & Label \\ \midrule
        John and Alice are planning a trip together. & John and Alice have finalized the details of their trip. & 1\\ \midrule
        Tom recently got a new job. & Tom successfully completed his first project at the new job. & 2 \\ \midrule
        Ellie did not enjoy her recent trip. & Ellie is looking forward to traveling again. & 3\\ \midrule
        Michael mentioned that he felt a lot of emotions on his wedding day. & Michael felt a lot of love from his family at the wedding. & 4\\ \bottomrule
    \end{tabular}%
    }
    \caption{Application of memory update actions from Table \ref{tab:memory_update_actions} in different context types }
    \label{table:memory_update_example} 
    \end{minipage}
\end{table*}

\newcommand{\thickhline}{\noalign{\hrule height 1pt}}
\begin{table*}[p]
    \centering
    \footnotesize
    \begin{tabularx}{\textwidth}{X}
        \thickhline
        \textbf{Instruction description} \\ \hline
        You are a conversation analyst tasked with examining two conversations from the movie \{moive\_name\}.
        Use your knowledge of the characters to enhance your analysis where applicable. \\ \\
        In your analysis, categorize the dialogue based on five criteria: \\ 
        1. \textbf{Persona Information}: Discuss aspects such as personality, job, age, education, favorite foods, music,
        hobbies, family life, daily activities, health, etc. \\
        2. \textbf{Temporal information}: Identify information that will soon become irrelevant, such as upcoming deadlines
        like "I need to submit my assignment by Friday" or temporary states like "I have a cold."\\
        3. \textbf{Shared Memory}: Focus on past experiences that the speakers refer to during their conversation, which
        they have previously experienced together. 
        This category includes both explicitly mentioned memories and those implied through their dialogue.\\
        For example, the exchange 'Alice: Wasn't that jazz festival we went to last summer amazing?' 'Bob: It was phenomenal, especially the live band under the stars.' should be
        categorized here because it indicates that Alice and Bob shared the experience of attending a jazz festival
        together. \\
        4. \textbf{Mutual Event}: This category captures significant events and interactions occurring directly between
        {speaker1} and {speaker2} during the current conversation, excluding any third-party involvement. Consider
        only those interactions that are substantial and directly involve both speakers. For example, from the
        exchange "Alice: Aren't these shoes pretty?", "Bob: Try them on.", "Alice: How do they look? Do they suit
        me?", you can extract that "Alice and Bob are experiencing shopping together."\\
        5. \textbf{None}: Assign this category to parts of the conversation that do not fit into the above categories. \\ \\
        Proceed to analyze the dialogue, addressing it one turn at a time:\\
        \{dialogues\_text\}\\
        Your task is to extract:\\
        - Persona information for \{speaker1\}\\
        - Persona information for \{speaker2\}\\
        - Temporal information for \{speaker1\}\\
        - Temporal information for \{speaker2\}\\
        - Shared memories between \{speaker1\} and \{speaker2\}\\
        - Mutual events occurring during the conversation between \{speaker1\} and \{speaker2\}\\ \\
        
        Format your findings by separating each category with  `***'. If no information is found for a category, indicate it with `None'. The expected format is:\\
        \texttt{[***Persona: \{speaker1\}'s information or `None'***Persona: \{speaker2\}'s information or `None'***Temporal: \{speaker1\}'s information or `None'***Temporal: \{speaker2\}'s information or `None'***Shared Memory: information or `None'***Mutual Event: information or `None'***]}\\
        Limit the output to 300 tokens to ensure concise and focused responses.\\
        For instance, the expected output should look like: \\
        \texttt[***Persona: Alice majors in artificial intelligence and enjoys pizza.***Persona: Bob is fond of hamsters.***Temporal: Alice has a medical check-up tomorrow.***Temporal: None***Shared Memory: Alice and Bob reminisce about attending a concert together.***Mutual Event: Alice and Bob are shopping together.***]\\ \\
        Present your responses directly, using the speakers' names without pronouns and avoiding category labels. 
        For instance, rather than stating "***Alice's temporal information includes an upcoming math project due tomorrow.***", simply note "***Temporal: Alice has a math project due tomorrow.***" \\ 
        Ensure that each analysis output is succinct, covering only the essential elements of the dialogue. Ensure you cover every part of the dialogue comprehensively. If a specific category does not apply, move on to the next without mention. Your detailed analysis will help illuminate the nuances of their interactions, capturing the essence of their shared and immediate experiences within the current dialogue.\\ \thickhline
    \end{tabularx}
    \caption{Prompt used for information extraction}
    \label{table:prompt_information_extraction}
\end{table*}

\begin{table*}[p]
    \centering
    \footnotesize
    \begin{tabularx}{\textwidth}{X}
        \thickhline
        \textbf{Instruction description} \\ \hline
        Please break down the following sentence into its core factual components without overly splitting the content.\\
        \{sentence\} \\
        For the output, list each cohesive factual unit with a number.\
        Ensure the breakdown retains natural phrasing while omitting any references to the significance, nature of the
        information, and discussions about the basis of any claims.\\
        Replace uncertain terms like "appears" or "seems" with more definitive expressions such as "is" to ensure the sentences convey clear and assertive information.\\
        Make sure to write in complete sentences and preserve the natural flow of information, excluding any explanations or justifications.\\
        \thickhline
    \end{tabularx}
    \caption{Prompt used for parsing information}
    \label{table:prompt_parsing_information}
\end{table*}

\begin{table*}[p]
    \centering
    \footnotesize
    \begin{tabularx}{\textwidth}{X}
        \thickhline
        \textbf{Instruction description} \\ \hline
        Labeling Task: As a dialogue analyst, your task is to systematically classify each line of dialogue from the
        provided conversation by linking it to detailed persona and context information. Evaluate each line of dialogue
        to determine which attributes from the provided persona and context information align best with the content.\\
        Dialogue Text:\\
        \{dialogues\_text\}\\
        Persona and Context Information:\\
        \{attr\_text\}\\
        Labeling Instructions:\\
        For each line of dialogue, identify and assign the most appropriate persona or context information available.\\
        Use "Everyday Language" only if the dialogue does not clearly fit any provided categories or pertains to routine conversation.\\
        General Output Guidelines: Assign labels that closely match the dialogue to relevant persona traits or contextual details. Strive to use each piece of information at least once to ensure comprehensive coverage of all provided attributes.\\
        Example of Expected Output:\\
        - Speaker1: "Do you know which artists are coming to the festival?"\\
        \textasteriskcentered Labels: Everyday Language\\
        - Speaker2: "Yes, BLACKPINK is coming to the festival! It's going to be amazing. Plus, there are lots of food trucks coming, including a skewer food truck which I'm happy about."\\
        \textasteriskcentered Labels: Speaker2 knows BLACKPINK is coming to the festival (Speaker2’s persona), Speaker2 likes skewers (Speaker2’s persona)\\
        - Speaker1: "Wow, that's awesome! We had so much fun at last year's festival, this year will be great too!" \\
        \textasteriskcentered Labels: Speaker1 and Speaker2 enjoyed last year's festival together (Shared memories)\\
        - Speaker2: "Totally looking forward to it!"\\
        \textasteriskcentered Labels: Speaker2 is looking forward to the festival (Speaker2’s temporal information)
        \\
        \thickhline
    \end{tabularx}
    \caption{Prompt used for labeling task}
    \label{table:prompt_labeling_task}
\end{table*}

\begin{table*}[p]
    \centering
    \footnotesize
    \begin{tabularx}{\textwidth}{X}
        \thickhline
        \textbf{Instruction description} \\ \hline
        You are a conversation analyst. \\
        You need to understand the context well and predict the next part of the dialogue.\\
        Based on the provided candidate memories and dialogue history, select all the appropriate memories for the next part of the conversation. \\
        These memories are elements that form the basis of the conversation.\\
        If no suitable memories are available, choose 'Everyday Language,' which refers to common, everyday expressions.\\ \\
        Task:\\
        Candidate Memories:\\
        \{candidates\}\\
        \\
        Dialogue History:\\
        \{dialogues\_text\} \\
        Select all the appropriate memories for the next part of the conversation by \{next\_speaker\}. \\
        If there are two or more memories, separate them with `\#\#\#':\\
        \thickhline
    \end{tabularx}
    \caption{Prompt used for memory selection}
    \label{table:prompt_memory_selection}
\end{table*}

\begin{table*}[p]
    \centering
    \footnotesize
    \begin{tabularx}{\textwidth}{X}
        \thickhline
        \textbf{Instruction description} \\ \hline
        Task: Generate the next response in a dialogue by focusing on the contextual cues detailed within parentheses in the dialogue history. Responses should be tailored according to the type of cue provided:\\ \\
        
        1. Memory-driven dialogues: If the cue within parentheses details specific character traits or background context, craft responses that reflect these memory-driven elements, ensuring character consistency and rich context.\\
        2. Everyday language dialogues: If the cue within parentheses is labeled "Everyday Language," generate responses that are based on typical day-to-day interactions, free from specific personas or detailed context.\\ \\
        
        \textbf{Dialogue History}:\\
        \{dia\_text\}
        \\ \thickhline
    \end{tabularx}
    \caption{Prompt used for response generation with memory}
    \label{table:prompt_response_generation}
\end{table*}

\begin{table*}[p]
    \centering
    \footnotesize
    \begin{tabularx}{\textwidth}{X}
        \thickhline
        \textbf{Instruction description} \\ \hline
        Task: Generate the next response in the dialogue based on the provided history. The response should logically follow and predict the next reply considering the context of the conversation.\\\\
        
        \textbf{Dialogue History}:\\
        \{dia\_text\}
        \\ \thickhline
    \end{tabularx}
    \caption{Prompt used for response generation without memory}
    \label{table:prompt_response_generation_without_memory}
\end{table*}

\begin{table*}[p]
    \centering
    \scriptsize
    \begin{tabularx}{\textwidth}{X}
        \thickhline
        \textbf{Instruction description} \\ \hline
        You are a language expert who understands the flow of conversation and manages memory.\\
        To effectively manage memory in a conversational system, it is crucial to understand the memory itself.\\
        As the conversation progresses, compare the information from previous sessions with the current session to update the memory and remove unnecessary sentences.\\
        Memory is categorized into the following four types:\\
        1. Persona information: This captures essential characteristics, including personality, occupation, and interests.\\
        2. Personal event: This information covers transient details like impending deadlines or current health conditions.\\
        3. Mutual event: This captures significant interactions between the speakers, focusing on substantial events directly involving both individuals. Over time, these mutual events become new shared memories.\\
        4. Shared memory: This refers to past experiences or memories that the two speakers have shared together prior to the current conversational context.\\
        Guidelines for Memory Management:\\
        Tasks to Perform in the Current Session:\\
        1. Remove incomplete information: Remove sentences that are incomplete or do not clearly convey the context.\\
            * Example: "SAM is interested in something." or "SAM mentions a place he visited."\\
        2. Remove information not suitable for conversation topics: Remove information that is irrelevant to the main topic of conversation.\\
            * Example: "JANE remembers SAM." or “JANE has a need to urinate.”\\
        3. Remove unrelated personal events: Remove personal event information that is not directly related to the individual or does not influence the conversation flow.\\
            * Example: "MARK talked about a coworker who went on vacation last month."\\
        4. Remove duplicate information: If the same information is provided in both Persona and Personal events, or if the same information is provided in Persona and Shared memory, remove the Persona and retain the other information.\\
            * Example: “KATE enjoys watching movies.” (Persona) and “KATE often watches movies on weekends.” (Personal event) provide similar information, so remove the Persona.\\
            * Example: “MIKE remembers the trip to Paris.” (Persona) and “MIKE and JANE shared a memorable trip to Paris.” (Shared memory) are similar; remove the Persona.\\
        5. Update Persona based on Mutual events: Update the Persona with emotions or reactions caused by Mutual events, and write sentences in the past tense.\\
            * Example: The Persona "JACK feels betrayed and angry." should be updated to "SARAH told JACK about her secret involvement in a rival project, causing JACK to feel betrayed and angry."\\
        Methods for Memory Update:\\
        1. Connect sequential/causal events: Link and update events that are sequential or have a causal relationship.\\
            * Example:\\
                * Previous memory:\\
                    * Tom recently got a new job.\\
                    * Tom was very nervous on his first day at work.\\
                * Current memory:\\
                    * Tom successfully completed his first project at the new job.\\
                * Updated memory:\\
                    * Tom recently got a new job and was very nervous on his first day. \\
                    * Tom has since successfully completed his first project.\\
        2. Update conflicting events: Reflect changes or transitions when the previous and current memories contain conflicting information.\\
            * Example:\\
                * Previous memory:\\
                    * Ellie did not enjoy her recent trip.\\
                    * Ellie said she would no longer plan trips.\\
                * Current memory:\\
                    * Ellie is planning a trip with her friends.\\
                    * Ellie is looking forward to traveling again.\\
                * Updated memory:\\
                    * Ellie did not enjoy her recent trip, but now she is planning a new trip with friends and is looking forward to it.\\
        3. Remove unnecessary personal event information: Exclude any unnecessary details about personal events. If the personal event only reflects a very short-term, trivial state (such as someone being in transit), it should be removed.\\
        	•	Example: "Jay is on the bus" should be removed.\\
        4. Accumulate unrelated events: Accumulate personal events that do not fit guidelines 1 through 3.\\
            * Example:\\
                * Previous memory:\\
                    * JANE likes spicy food.\\
                * Current memory:\\
                    * JANE dislikes math.\\
                * Updated memory:\\
                    * JANE likes spicy food.\\
                    * JANE dislikes math.\\
        5. Use the past tense for Mutual events: Mutual events from the current session become past events, so convert them to the past tense.\\
            * Example:\\
                * Previous memory:\\
                    * John and Alice are planning a trip together.\\
                * Current memory:\\
                    * John and Alice have finalized the details of their trip.\\
                * Updated memory:\\
                    * John and Alice planned a trip together and have finalized the details.\\
        Actual Content Update:\\
        Use the following structure to update the memory based on the provided guidelines.\\
        All sentences in the updated memory must start with a person’s name.
        Previous memory:\\
        \{previous\_memory\}\\
        Mutual event:\\
        Current memory:\\
        \{current\_memory\}\\
        Updated memory:\\
        \thickhline
    \end{tabularx}
    \caption{Prompt used for memory update}
    \label{table:prompt_memory_update}
\end{table*}

\begin{table*}[p]
    \centering
    \footnotesize
    \begin{tabularx}{\textwidth}{X}
        \thickhline
        \textbf{Instruction description} \\ \hline
        You will be given a conversation between two participants. Your task is to read, remember, and understand the dialogue to evaluate how logically consistent and naturally flowing the interaction is, reflecting a coherent and human-like conversation.\\
        Coherence Evaluation Criteria: Coherence evaluates how logically consistent, human-like, and naturally flowing a conversation is. It measures the natural flow of ideas and the relevance of responses to previous statements. \\
        Even if the conversation shifts topics, it remains coherent if the shifts are contextually appropriate. Emotional depth, such as nervousness or abruptness, should not automatically be seen as incoherence; instead, it can reflect the human aspect of the dialogue. \\\\
        Coherence is further enhanced by meaningful narrative progression and the appropriate use of figurative language or intentional topic shifts that align with the participants' intent.\\ 
        Sudden topic changes, especially when discussing personal concerns or emotions, can still feel natural and coherent if the shifts are handled smoothly, keeping the flow of the conversation intact. 
        Recognizing emotions like nervousness or frustration can contribute to coherence, as these reactions add depth and realism to the interaction. 
        Thus, human-like reactions and shifts in tone should be considered part of a coherent conversation. \\ \\
        Key Aspects to Consider:\\
        Engaging Narrative and Logical Progression: The conversation should maintain a logically consistent and connected narrative, even when the topic shifts or tone changes. Look for ways the dialogue remains relevant and engaging through transitions between ideas. Multiple topic shifts can still be coherent if each shift matches or connects well with the previous topic, maintaining the conversation's flow.\\
        Adaptability and Smooth Flow: Shifts in topic or tone should feel connected. If there is tension or disagreement, consider how it adds depth and coherence to the dialogue, contributing to its narrative progression. The flow should feel authentic, with responses adding to the logical continuity of the dialogue. \\
        Logical Response Relevance: Each response should directly address or logically follow from the previous statement or question. A coherent conversation maintains a clear connection between responses, avoiding irrelevant or off-topic answers. When participants introduce new ideas or shift topics, these should still feel grounded in the context of the prior dialogue, ensuring that the conversation remains connected and purposeful. \\ \\
        Scoring Guidelines:\\
        0 points: The conversation lacks coherence. Responses are disjointed, irrelevant, or nonsensical, making the dialogue feel artificial and disconnected.\\
        1 point: The conversation shows minimal coherence. Responses are occasionally relevant but often feel awkward or forced, disrupting the natural flow of dialogue.\\
        2 points: The conversation is moderately coherent. Most responses make sense and follow the context, contributing to a relatively smooth and connected dialogue, even when dealing with intense or emotionally charged topics.\\
        3 points: The conversation is highly coherent, with responses that flow naturally, maintain logical progression, and closely resemble a human-like conversation. Shifts in tone or topic, even if intense or emotionally charged, should contribute positively to the overall narrative, maintaining a seamless and dynamic flow.\\\\
        The output format should be as follows:\\
        Score: [score]\\
        Now, based on the provided conversation, evaluate which response has better Coherence considering the flow, logical consistency, and emotional depth of the dialogue.\\
        Dialogue : \\
        \{dialogue\}\\
        Score :\\
        \thickhline
    \end{tabularx}
    \caption{Prompt used for coherence evaluation}
    \label{table:prompt_multisession_coherence}
\end{table*}

\begin{table*}[p]
    \centering
    \footnotesize
    \begin{tabularx}{\textwidth}{X}
        \thickhline
        \textbf{Instruction description} \\ \hline
        You will be given a conversation between two participants. Your task is to read, understand, and evaluate the interaction based on how it subtly reveals emotional complexity, relational depth, and captivating dynamics. The conversation should reflect moments of emotional engagement, relational tension, or even quiet significance that makes the interaction feel engaging and memorable.\\
        Engagingness Evaluation Criteria:
        Engagingness measures how much the conversation entertains through emotional layers, subtle relational tension, or significant moments that might not immediately appear dramatic but still add depth. The conversation should feel dynamic and engaging, with interactions that reveal or suggest deeper emotional connections or personal significance, even in seemingly calm exchanges. It's important to consider how elements of trust, vulnerability, or tension might quietly enrich the dialogue, making it captivating and entertaining.\\  \\
        Key Aspects to Consider:\\
        Subtle Emotional Depth and Relational Tension: The conversation should contain moments where emotional layers or relational tension are hinted at, even if not overt. Subtle expressions of humor, trust, or tension can add richness to the interaction, enhancing engagement through underlying dynamics between participants.\\
        Significant Moments, Even in Calm Exchanges: The conversation may include moments that seem quiet or routine on the surface but carry significant emotional weight or meaning. These moments should be considered memorable, as they contribute to the deeper dynamics of the interaction, leaving a lasting impression.\\
        Consistent Engagement and Entertainment Value: The conversation should create a flow that keeps the reader invested, revealing dynamics between participants that make the interaction entertaining, even in subtle ways. The dialogue may include calm exchanges, but it should feel engaging through the emotions and relational depth that lie beneath the surface.\\\\
        Scoring Guide:\\
        0 points: The conversation lacks emotional complexity, depth, or relational engagement, making it trivial or forgettable.\\
        1 point: The conversation has minimal emotional engagement, with few moments of relational tension or humor, but they are not sustained or impactful.\\
        2 points: The conversation includes moments of emotional engagement or humor but is inconsistent in creating a strong impact throughout.\\
        3 points: The conversation consistently engages and entertains through emotional layers, relational depth, or significant moments that reveal underlying dynamics. Even seemingly simple or calm exchanges carry emotional weight, making the interaction feel rich and captivating, with memorable moments that enhance the overall engagement.\\\\
        The output format should be as follows:\\
        Score: [score]\\
        Now, based on the conversation below, evaluate the Engagingness score by considering how captivating, creative, and emotionally resonant the dialogue is.\\
        Dialogue : \\
        \{dialogue\}\\
        Score :\\
        \thickhline
    \end{tabularx}
    \caption{Prompt used for engagingness evaluation}
    \label{table:prompt_multisession_engagingness}
\end{table*}

\begin{table*}[p]
    \centering
    \footnotesize
    \begin{tabularx}{\textwidth}{X}
        \thickhline
        \textbf{Instruction description} \\ \hline
        You will be given a conversation between two participants. Your goal is to read, remember, and understand the dialogue to evaluate how well it reflects the participants' mutual understanding and familiarity with each other.\\
        
        Closeness Evaluation Criteria:\\
        
        Closeness measures how well the participants know each other, reflecting the depth of their relationship and the extent of their understanding of one another. This evaluation goes beyond mere communication, assessing how their interactions reveal a shared history, familiarity, or dynamic that shapes their connection. Whether the interaction is friendly, contentious, or competitive, the key is how well the participants recognize and respond to each other's traits, emotions, and communication styles. Elements such as light teasing, the use of slang, emotional support, sharing personal information, forming empathy, and communication flexibility are vital indicators of closeness, even in adversarial exchanges.\\
        \\
        Key Aspects to Consider:\\
        
        Depth of Mutual Recognition: Look for moments where the participants acknowledge and respond to each other's unique traits and past experiences, including light-hearted jokes or teasing, indicating familiarity and understanding.\\
        
        Emotional Support and Personal Sharing: Assess how the participants provide emotional support during challenging times and share personal information, such as discussing family matters or past mistakes, which enhances their connection.\\
        
        Adaptation to Each Other's Responses: Evaluate how the participants adjust their behavior or language in response to one another, showcasing their comfort and familiarity. This adaptation can reflect a complex relationship where they navigate each other's emotions effectively.\\
        \\
        Scoring Guide:\\
        
        0 points: The conversation lacks any sense of closeness. There is no mutual understanding or familiarity, making the interaction feel impersonal and disconnected.\\
        
        1 point: The conversation shows minimal closeness. There are hints of mutual understanding, but the interaction remains largely superficial and lacks depth.\\
        
        2 points: The conversation reflects moderate closeness. The participants demonstrate a reasonable understanding of each other through their exchanges, recognizing each other's traits, motivations, and emotional states.\\
        
        3 points: The conversation demonstrates high closeness. The interaction reveals a deep familiarity and consistent recognition of each other's character, history, and dynamics. The dialogue feels like a natural extension of a long-standing relationship, reflecting profound mutual understanding, whether the context is friendly, competitive, or contentious.\\
        \\
        The output format should be as follows:\\
        
        Score: [score]\\
        
        Now, based on the provided conversation, evaluate which response demonstrates better Closeness, focusing on how well the participants understand each other, reflect familiarity, and adapt to each other's traits, emotions, and communication styles.\\
        
        Dialogue :\\
        \{dialogue\}\\
        Past memory :\\
        \{shared\_memory\}\\
        Score:\\
        \thickhline
    \end{tabularx}
    \caption{Prompt used for closeness evaluation}
    \label{table:prompt_multisession_closeness}
\end{table*}

\begin{table*}[p]
    \centering
    \footnotesize
    \begin{tabularx}{\textwidth}{X}
        \thickhline
        \textbf{Instruction description} \\ \hline
        You will be given a conversation between two participants. Your task is to read, remember, and understand the dialogue to evaluate how well it reflects the participants' shared memories, past interactions, and mutual history.\\
        
        Reflectiveness Evaluation Criteria: Reflectiveness measures how well the conversation incorporates shared experiences, past events, and personal history between the participants. It evaluates how memories and mutual understanding are used to inform the interaction, reflecting a deeper connection that goes beyond surface-level dialogue. The primary focus is on the integration of past experiences and how well these memories are acknowledged, referenced, or built upon throughout the conversation.\\
        \\
        Key Aspects to Consider:\\
        
        Integration of Shared Memories: Assess how well the participants incorporate past events, memories, or experiences into the conversation. This can include direct references to shared history, implicit acknowledgments of past interactions, or the subtle use of memories that add depth and context to the dialogue.
        \\
        Consistency with Past Interactions: Evaluate how consistently the conversation aligns with previously established dynamics, events, or shared experiences. The interaction should feel like a continuation of their mutual history, with responses that logically and emotionally connect to their past.\\
        
        Mutual Recognition of History and Context: Look at how the participants recognize and respond to the shared context of their relationship. This includes acknowledging each other's past actions, decisions, or shared journeys, demonstrating that their connection is informed by a rich and evolving history.\\\\
        
        Scoring Guide:\\
        
        0 points: The conversation lacks any reflectiveness. There is no sense of shared history or mutual recognition of past experiences, making the dialogue feel disconnected from their relationship.\\
        
        1 point: The conversation shows minimal reflectiveness. There are some hints of shared memories or past events, but the interaction lacks depth and consistency in incorporating these elements.\\
        
        2 points: The conversation reflects moderate reflectiveness. Shared memories and past experiences are reasonably integrated, adding context and depth to the interaction, even if subtle.\\
        
        3 points: The conversation demonstrates high reflectiveness. The interaction consistently incorporates shared history and past interactions, weaving memories into the dialogue in a meaningful and engaging way that enriches their connection.\\
        \\
        The output format should be as follows:\\
        
        Score: [score]\\
        
        Now, based on the provided conversation, evaluate the Reflectiveness score by considering how well the dialogue incorporates shared memories, past events, and the mutual history between the participants.\\
        
        Dialogue :\\
        \{dialogue\}\\
        Past memory :\\
        \{shared\_memory\}\\
        Score:\\
        \thickhline
    \end{tabularx}
    \caption{Prompt used for reflectiveness evaluation}
    \label{table:prompt_multisession_reflectiveness}
\end{table*}

\begin{table*}[p]
    \centering
    \footnotesize
    \begin{tabularx}{\textwidth}{X}
        \thickhline
        \textbf{Instruction description} \\ \hline
        Your task is to evaluate the "Consistency" between the following five consecutive conversations. Read through all the conversations carefully and assess whether the relationship between the participants remains consistent throughout the dialogue. Pay particular attention to how shifts in tone or behavior are well-supported by the context or emotional developments, allowing the relationship to evolve naturally. Even if the dialogue features notable changes in the characters' tone or attitude, these shifts should still be seen as consistent if they are explained or connected to previous conversations.\\
        Evaluation Criteria:\\
        Consistency measures the evolving nature of the relationship between the participants across the conversations. Conversations with good consistency will feel natural, even if the relationship experiences shifts in tone or intensity, as long as these shifts are well-supported by prior interactions or context. Shifts that show character growth or new revelations, and that maintain logical progression based on the overall relationship dynamics, should contribute positively to the consistency score.\\
        0 Points: The conversations lack consistency, with abrupt shifts in tone or behavior that feel unsupported or out of context.\\
        1 Point: The conversations show some consistency, but the relationship between the participants experiences a few unexplained or awkward shifts.\\
        2 Points: The conversations have moderate consistency. The relationship evolves in a mostly coherent manner, though there may be minor shifts that slightly disrupt the flow.\\
        3 Points: The conversations exhibit excellent consistency, with all shifts in tone or behavior feeling logical and well-supported by prior dialogue. In these conversations, even as the tone or intensity changes, these shifts are always justified by emotional or contextual developments between the characters. The relationship evolves naturally and feels enriched by the deeper emotional understanding of both participants. Even if the characters' actions or speech become more direct or intense, the progression makes sense within the established dynamics of their relationship.\\
        Output Format:\\
        Score : [score]\\
        Dialogue 1 :\\
        \{dialogue1\}\\
        Dialogue 2 :\\
        \{dialogue2\}\\
        Dialogue 3 :\\
        \{dialogue3\}\\
        Dialogue 4 :\\
        \{dialogue4\}\\
        Dialogue 5 :\\
        \{dialogue5\}\\
        Score :\\
        \thickhline
    \end{tabularx}
    \caption{Consistency prompt for Episode evaluation.}
    \label{table:prompt_episode_consistency}
\end{table*}

\begin{table*}[p]
    \centering
    \footnotesize
    \begin{tabularx}{\textwidth}{X}
        \thickhline
        \textbf{Instruction description} \\ \hline
        Your task is to evaluate the "Reflectiveness" between the participants across the following five consecutive conversations. Reflectiveness measures how well the relationship between the participants can be inferred from their dialogue. Focus on whether the conversations naturally reveal their connection, providing clear and consistent clues about how they relate to each other.\\
        Evaluation Criteria:\\
        Reflectiveness measures how well the participants' relationship can be inferred from their dialogue. It is rated higher when conversations provide clear, consistent clues about their connection, such as personal interactions, emotional exchanges, or shared experiences that suggest familiarity or trust.\\
        Reflectiveness is rated lower when the relationship is ambiguous, or when the dialogue lacks personal details, shared experiences, or emotional exchanges. \\Conversations that focus on surface-level information or general topics without providing insights into the participants' bond make it difficult to infer their relationship. If the dialogue remains formal or distant, without revealing any personal or emotional depth, the score decreases.\\
        Scoring Criteria:\\
        0 points: The conversations provide no clear indications of the participants' relationship, leaving their connection undefined.\\
        1 point: The conversations provide limited or vague information, making it hard to infer the participants' relationship.\\
        2 points: The conversations give some insight into the relationship but lack strong, consistent cues. There is a moderate sense of connection, but not enough clarity about their bond.\\
        3 points: The conversations clearly and consistently reveal the participants' relationship. Their interactions, whether direct or subtle, provide consistent cues about trust, emotional engagement, or shared history. These exchanges make it easy to define their relationship, with personal details, familiarity, or mutual understanding naturally emerging throughout the conversations.\\
        Output Format:\\
        Score : [score]\\
        Dialogue 1 :\\
        \{dialogue1\}\\
        Dialogue 2 :\\
        \{dialogue2\}\\
        Dialogue 3 :\\
        \{dialogue3\}\\
        Dialogue 4 :\\
        \{dialogue4\}\\
        Dialogue 5 :\\
        \{dialogue5\}\\
        Score :\\
        \thickhline
    \end{tabularx}
    \caption{Reflectiveness prompt for Episode evaluation.}
    \label{table:prompt_episode_reflectiveness}
\end{table*}

\begin{table*}[p]
    \centering
    \footnotesize
    \begin{tabularx}{\textwidth}{X}
        \thickhline
        \textbf{Instruction description} \\ \hline
        Your task is to evaluate the "Engagingness" between the participants in the following five consecutive conversations. Read through all the conversations carefully and assess how entertaining and engaging the dialogue is for the participants. Pay particular attention to whether the conversations are connected across sessions and if this connection enhances the overall fun and interest of the dialogue. A clear and engaging flow between sessions can increase the overall Engagingness of the conversation.\\
        Evaluation Criteria:\\
        Engagingness measures how fun and captivating the conversations are. High levels of engagingness will make the dialogue feel lively, entertaining, and enjoyable for both participants. Additionally, conversations that show a strong connection across sessions, enhancing the overall flow and enjoyment, will have higher levels of engagingness. When the participants talk about shared experiences or events, it helps them understand each other better, which overall makes the conversation more enjoyable and engaging.\\
        0 points: The conversations are not engaging. The interactions are dull, monotonous, and fail to capture any interest. There is no meaningful connection between sessions.\\
        1 point: The conversations show minimal engagingness. There are a few moments of interest, but the overall dialogue feels flat and lacks energy. The connection between sessions is weak or nonexistent.\\
        2 points: The conversations are moderately engaging. The participants manage to create some entertaining exchanges, but the overall flow isn't consistently fun or lively. Connections between sessions exist but are not particularly strong or noticeable.\\
        3 points: The conversations are highly engaging, with dynamic, fun, and captivating exchanges that keep the conversation entertaining and full of energy throughout.\\
        The dialogue flows well between sessions, and this connection makes the overall experience more enjoyable. When participants share common experiences, it helps them know each other better, making the conversation feel even more fun and engaging.\\
        Output Format:\\
        Score : [score]\\
        Dialogue 1 :\\
        \{dialogue1\}\\
        Dialogue 2 :\\
        \{dialogue2\}\\
        Dialogue 3 :\\
        \{dialogue3\}\\
        Dialogue 4 :\\
        \{dialogue4\}\\
        Dialogue 5 :\\
        \{dialogue5\}\\
        Score :\\
        \thickhline
    \end{tabularx}
    \caption{Engagingness prompt for Episode evaluation.}
    \label{table:prompt_episode_engagingness}\
\end{table*}
\captionsetup[figure]{justification=raggedright,singlelinecheck=false}

\captionsetup[figure]{justification=raggedright,singlelinecheck=false}

\end{document}